\documentclass{article}
\usepackage[preprint]{corl_2026}
\usepackage{amsmath,amssymb,booktabs,multirow,graphicx,array}
\usepackage{xcolor}
\usepackage{url}
\usepackage{hyperref}
\hypersetup{
    pdftitle={When Will You Hand It Over? A Hidden Start Delay Makes Self-Play Handover Robust to Unfamiliar Partners},
    pdfauthor={Mohammad Khoshnazar, Mike Degany, Deyuan Qu, Qing Yang, Matyas Szanto},
    pdfsubject={Accepted at the HumanRLI Workshop, CoRL 2026},
    pdfkeywords={Human-Robot Handover, Reinforcement Learning, Self-Play, Partner Generalization}
}
\newcommand{\ours}{\textsc{CoPlay-Handover}}
\newcommand{\hsd}{\textsc{HSD}}

\newcommand{\robot}{\mathrm{R}}
\newcommand{\human}{\mathrm{H}}

\title{When Will You Hand It Over?\\A Hidden Start Delay Makes Self-Play Handover\\Robust to Unfamiliar Partners}
\author{
Mohammad Khoshnazar$^{1}$ \quad
Mike Degany$^{2}$ \quad
Deyuan Qu$^{3}$ \quad
Qing Yang$^{2}$ \quad
Matyas Szanto$^{4}$\\[0.5em]
$^{1}$Universit\"at Bremen \quad $^{2}$University of North Texas\\
$^{3}$Toyota Motor North America \quad
$^{4}$Budapest University of Technology and Economics
}
\begin{document}
\maketitle
\begin{center}
\small
\textbf{Code:} \url{https://github.com/mkhoshnam/Selfplay-PPO-handover}
\end{center}
\begin{abstract}
Cooperative self-play can teach a humanoid to receive an object without any human data, but the partner it meets after training is never the one it trained with; we show that a fluent self-play handover policy breaks with unfamiliar partners, identify the main cause, and repair it. In a physics-based parcel handover between a Unitree G1 and a learned giver, both trained by alternating PPO self-play, the receiver completes 95\% and 90\% of handovers with its own partner (two training seeds) but only 31\% with scripted partners it has never met. Paired interventions isolate the cause: under a shared reward the learned giver always starts its offer at once, so the receiver learns a single rendezvous, and delaying its own partner by one second cuts success from 95\% to 6\% (87\% to 34\% in the second seed). We introduce hidden start-delay randomisation (\hsd{}), a one-line rule on the learned partner that the receiver never observes. After 300 further updates, success with unseen scripted partners rises from 31\% to 80\% in both seeds, against 35--41\% for a budget-matched control. The gain holds on a fresh scenario seed, on eleven held-out partner conditions and on partners that start later than any training delay; it cuts high-force contact three- to five-fold and needs no retraining of the partner.
\end{abstract}

\begin{figure}[h!]
\centering
\includegraphics[width=\linewidth]{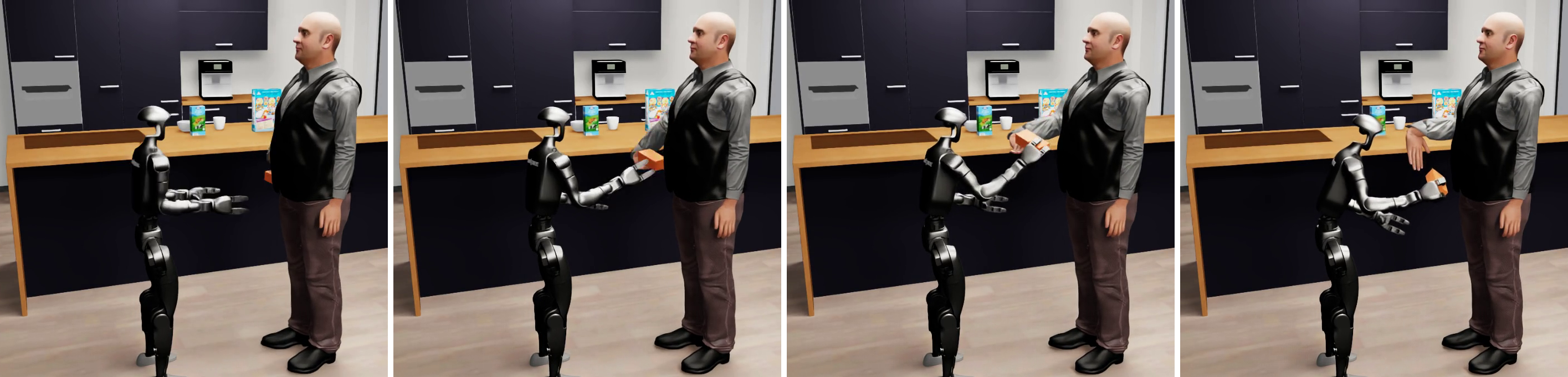}
\caption{\textbf{The task.} A Unitree G1 with a Dex3 hand receives a parcel from a learned partner: it reaches, shares the load, and holds the parcel alone once the partner lets go (one episode of the self-play policy with its co-trained partner). The character is an animation driven by the partner's hand, and the apartment is scenery only.}
\label{fig:teaser}
\end{figure}

\section{Introduction}
A handover is a joint action. Giver and receiver must agree not only on where the object changes hands but on \emph{when}: the receiver has to arrive, take the weight and keep the object supported while the giver lets go, and each side times its part on the other \citep{strabala2013towards,ortenzi2021handovers}. A robot that receives objects from people therefore faces givers who start when they are ready, not when the robot expects them to: a human-induced distribution shift that a learned policy must absorb.

Cooperative self-play is an attractive way to learn this coordination, because it needs no demonstrations: a receiver and a giver are trained together and discover a fluent exchange. But a policy pair that is fluent with itself may have agreed on a private convention, a failure known from cooperative games \citep{hu2020otherplay,carroll2019utility}. Whether such conventions arise in contact-rich physical handover, what they consist of, and how to remove them has not been established. We study these questions in a physics-based parcel handover between a fixed-base Unitree G1 and a learned, human-constrained partner (Figure~\ref{fig:teaser}).

The convention exists, and it is temporal. The self-play receiver succeeds in 95.0 / 89.5\% of handovers with its own partner (two training seeds) but in only 30.5 / 30.5\% with scripted partners it has never met. We trace this collapse mainly to one variable, the moment the partner starts. Because the partner shares the receiver's reward, it always starts its offer immediately, and the receiver learns to take the parcel on the fly at a fixed moment of the partner's reach. Delaying the receiver's own partner by one second, with everything else unchanged, reproduces the failure (95\% to 6\%, and 87\% to 34\%); making a scripted partner prompt and smooth removes most of it.

The diagnosis yields a simple and effective repair. Hidden start-delay randomisation (\hsd{}) keeps the learned partner's hand at rest for a hidden, per-episode time before it may begin to offer; the receiver's observations, rewards and network are untouched. After 300 further updates the receiver reaches 80.5 / 80.0\% with unseen scripted partners, 2.6 times its previous level, while a budget-matched continuation without \hsd{} stays at 35.0 / 41.0\%. The robustness extends to a fresh scenario seed, to eleven held-out partner conditions and to partners that start later than any delay seen in training.

\paragraph{Contributions.} (i) \ours{}, a physically coupled two-policy handover framework in which the giver's offer and release are learned, together with a support-aware success measure (Figure~\ref{fig:overview}). (ii) A causal diagnosis by paired interventions: the failure of self-play handover with unfamiliar partners is caused mainly by partner start timing, not by partner identity or by the partner's fine corrections. (iii) \hsd{}, a hidden constraint on the partner that repairs the failure, validated against a matched control, on a fresh scenario seed, on held-out conditions and with a frozen partner. (iv) A second hidden rule, on release, that teaches the receiver to hold still for partners that require it.

\section{Related Work}
\paragraph{Robot handover.} Object handover involves spatial approach, contact establishment, force sharing and release timing \citep{strabala2013towards,ortenzi2021handovers,duan2024handover}. Prior work investigates adaptive human--robot handover coordination \citep{huang2015adaptive,wu2022adaptive} and sensory feedback for fluent release \citep{penzotti2025release}. Simulation benchmarks for learned human-to-robot handover build on recorded human motion \citep{chao2022handoversim}, where the giver follows a prepared trajectory. We instead make the giver's offer and release the outputs of a learned policy and study the physical exchange itself: who carries the weight, and when the giver lets go.

\paragraph{Learning with changing partners.} Multi-agent actor--critic learning and self-play model agents that adapt to one another \citep{lowe2017maddpg,carroll2019utility}. Self-play agents settle on narrow conventions that fail with new partners \citep{hu2020otherplay}; population approaches train against historical or independently trained partners \citep{lanctot2017psro,jaderberg2019human,strouse2021fcp}, although policy diversity need not imply diverse strategies \citep{biswas2024pitfalls}. Hidden-utility self-play diversifies partners through hidden reward biases \citep{yu2023hsp}, and Co-GAIL learns diverse collaboration strategies, including handover, from demonstrations \citep{wang2022cogail}. We show the convention failure in a contact-rich physical exchange, measure its cause, and show that a snapshot pool does not prevent it, whereas a hidden hard constraint on one physical variable does.

\paragraph{Robust policy adaptation.} Recent work studies robustness to temporal mismatch in generated future guidance \citep{khoshnazar2026rafc} and to changes in robot actuator capabilities \citep{khoshnazar2026capable}. These address changes in the policy's guidance or embodiment; our work instead studies changing partner behavior during a physically coupled handover.

\section{Task and Success Measure}
\subsection{Coupled handover dynamics}
At each control step the robot receives an observation $o_t^{\robot}$ and the partner its own observation history $o_t^{\human}$, and two policies generate actions,
\begin{equation}
 a_t^{\robot}\sim\pi_{\theta_{\robot}}(\cdot\mid o_{\le t}^{\robot}),\qquad
 a_t^{\human}\sim\pi_{\theta_{\human}}(\cdot\mid o_t^{\human}).
\end{equation}
The robot commands ten waist and arm joint-position targets and one hand closure. The partner controls how fast it reaches (it can slow down, pause or back off), small corrections of the offer position ($\pm2$\,cm) and wrist tilt, and a release request. The request takes effect only under human-plausible conditions: the robot's hand touches the parcel, the partner feels less than 80\% of the load, and its reach is at least 60\% complete.

The simulation runs in Isaac Lab with Isaac Sim/PhysX \citep{mittal2023orbit} at a 30\,Hz control rate, with episodes of up to 8\,s. The Dex3 hand interacts with a rigid 0.5\,kg parcel ($7\times16\times5$\,cm) through articulated finger contact. The giver's hand is a physical body whose trajectory obeys speed (1\,m/s), acceleration (5\,m/s$^2$) and turn-rate limits; the offer point, rest point and nominal tilt are drawn per episode. The human-like character in our figures is an animation driven by this hand, so we say \emph{partner}, not \emph{human}.

\subsection{What each actor observes}
Neither actor consumes an image. The receiver's 89-dimensional observation contains its arm and finger joint state, palm, parcel and partner-hand poses, the error to the grasp pose, tactile forces, a thumb--finger opposition flag, the wrist force and its previous action. The partner's 26-dimensional observation contains its own hand pose, the robot palm and the parcel relative to its hand, the load it feels, its reach and release state, and whether the robot's hand touches the parcel and how far it has closed: what a giver can see and feel. Neither actor reads the other's internal command, and the receiver never observes any hidden partner rule. Poses come from simulation, without a perception module.

\subsection{Supported transfer}
A handover in which the parcel is dropped and caught, or the robot collides with the giver's hand, should not count as a success. We call a control step \emph{unsupported} when the partner's grip and the robot's hand together carry less than half the parcel weight:
\begin{equation}
 u_t=\mathbb{1}\!\left[|F_{z,t}^{H}|+|F_{z,t}^{R}|<\tfrac12 mg\right].
 \label{eq:unsupported}
\end{equation}
Let $A$ be the acquisition event: after the partner's hand has opened, the robot alone holds the parcel with a formed grip for 15 consecutive control steps. Our success criterion, \emph{supported transfer}, additionally demands that the whole interaction was physically sound:
\begin{equation}
\mathrm{ST}=\mathbb{1}[A]\;\mathbb{1}\!\Big[\sum_{t}u_t\le 3\Big]\;\mathbb{1}\big[\max_t F^{\robot\human}_t\le 20\,\mathrm{N}\big]\;\mathbb{1}\big[\text{slip}\le 3\,\mathrm{cm}\big],
\label{eq:supported}
\end{equation}
where the sum runs over the entire episode, $F^{\robot\human}$ is the contact force between the robot and the partner's hand, and slip is the parcel's travel inside the robot's grasp. All percentages in this paper are supported transfer.

\begin{figure}[t]
\centering
\includegraphics[width=\linewidth]{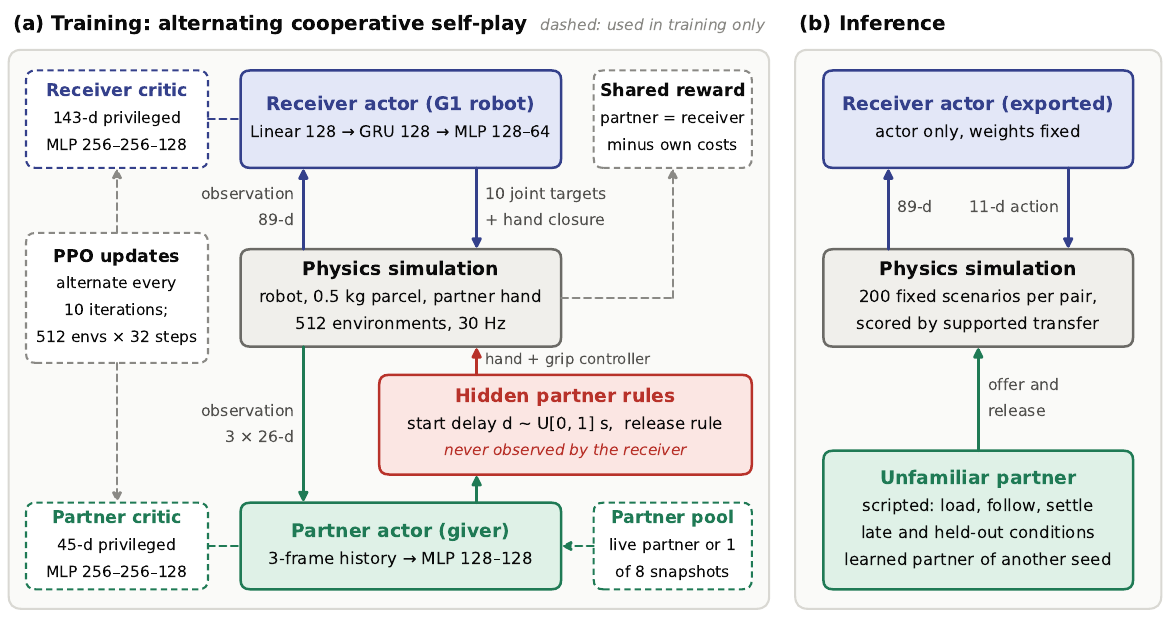}
\caption{\textbf{\ours{}.} (a) Training: receiver and partner act in the same physics simulation and are updated in alternation by PPO, each with a privileged critic. Hidden partner rules sit between the partner's action and its hand controller and are never observed by the receiver. (b) Inference: only the receiver actor is kept, and it is evaluated with partners it has never met.}
\label{fig:overview}
\end{figure}

\section{Method}
Figure~\ref{fig:overview} summarises the training and inference paths.

\subsection{Alternating cooperative self-play}
The receiver actor is a recurrent (GRU) network and the partner actor an MLP over a three-frame history; each agent has its own observation normaliser and a privileged critic that is used only in training \citep{pinto2017asymmetric}. Both actors act at every step, but only one actor--critic pair is updated at a time, alternating every 10 iterations (512 environments, 32 steps per iteration, $\gamma=0.99$, $\lambda=0.95$). Each agent computes generalised advantage estimates from its own rewards and optimises the PPO clipped-surrogate objective \citep{schulman2016gae,schulman2017ppo}. A four-stage curriculum leads from ``parcel between the fingers'' to the full reach and partner range.

The robot reward combines progress toward the grasp pose, load sharing, a formed grasp, holding and completion, with costs for unsupported flight, slip, impact, robot--partner contact and drops. The partner's reward is
\begin{equation}
 r_t^{H}=r_t^{R}-c_{\rm imp}C_t- c_{\rm rel}P_t-c_{\rm nat}N_t-c_{\rm act}\Vert a_t^H-a_{t-1}^H\Vert_2^2,
 \label{eq:reward}
\end{equation}
where $C_t$ indicates continued carrying, $P_t$ flags a release before the receiver has a formed grip under load, and $N_t$ measures deviation from a natural offering motion ($c_{\rm imp}=0.25$, $c_{\rm rel}=40$, $c_{\rm nat}=2$, $c_{\rm act}=0.01$). The partner shares the robot's reward, so anything that delays completion costs it.

A partner pool retains up to eight frozen snapshots of the partner actor, one added every 50 iterations; each episode uses the live partner or a snapshot with probability 0.5.

\subsection{Hidden partner rules}
\label{sec:release}
A shared reward removes any incentive for the partner to vary, and, as Section~\ref{sec:convention} shows, a pool of past partners does not restore it. We therefore impose variety directly, as hidden, hard, per-episode rules on the learned partner, and let the partner learn everything else.

\paragraph{Hidden start delay (\hsd{}).} At reset a delay $d\sim\mathcal U[0,d_{\max}]$ is drawn, with $d_{\max}=1$\,s. While $t<d$ the partner's reach rate is forced to zero: its hand stays at rest holding the parcel, and it is not charged for the enforced wait. $d_{\max}$ is ramped from zero over 150 iterations. The rule is one line in the partner's controller and changes no observation, reward or network; the receiver must infer from the partner's motion when the offer begins. \hsd{} is deliberately minimal: it randomises the one variable singled out by the diagnosis, and the matched control attributes the gain to it.

\paragraph{Hidden release rule.} Each episode the partner is assigned a hidden style, and its release request takes effect only once the style's condition holds: \emph{load}, as before; \emph{follow}, its hand yields to the pull and the parcel has been drawn 3--6\,cm; \emph{settle}, the receiver carries at least half the weight and has held still for 0.4--0.8\,s.

\section{Experimental Protocol}
\paragraph{Runs and controls.} The reference, \emph{self-play only}, trains receiver and partner from scratch for 800 updates each (13.1 million robot samples) with the pool. Every intervention is a matched continuation of this reference (300 receiver and 300 partner updates, 4.9 million robot samples), so that the arms differ only in the intervention: (i) \emph{control}: continue unchanged; (ii) \emph{\hsd{}}; (iii) \emph{\hsd{}, frozen partner}: the identical schedule with the partner never updated; (iv) \emph{\hsd{}+release rule}: continue (ii) with the release rule. Every run uses two training seeds, reported as ``seed 42 / seed 43''.

\paragraph{Partners.} Each receiver is tested with (a) its own learned partner, with and without start delay; (b) the learned partner of the other seed; and (c) scripted partners that never appear in training. The scripted partner draws its start time (0.2--1.2\,s), reach duration, hesitation, sway and release thresholds per episode (\texttt{train}) and has three release families, \emph{load}, \emph{follow} and \emph{settle}; the scripted settle partner additionally withdraws for good if the parcel is displaced more than 3\,cm before it lets go, and is treated in Section~\ref{sec:families}. An eleven-condition held-out bank varies the partner beyond these ranges; its \texttt{late} condition starts later than any training delay.

\paragraph{Evaluation.} Every receiver--partner cell uses the same fixed table of 200 scenarios (100 per condition for the held-out bank). The self-play, \hsd{} and frozen-partner arms are re-evaluated on a \emph{fresh} scenario seed that was never used for diagnosis, design or model selection, and the equivalence margin of the frozen-partner comparison (five points) was fixed in writing before that run. With 200 scenarios, the 95\% binomial interval of a cell is at most $\pm7$ points. Contact simulation is not bit-reproducible: repeated evaluations of the same pair differ by up to 2.5 points.

\section{Results}
\subsection{Self-play learns a fluent handover and a single timing convention}
\label{sec:convention}
Alternating self-play produces a coordinated, physically sound handover: the receiver achieves 95.0 / 89.5\% supported transfer with its co-trained partner, and 96.0 / 79.5\% with the partner of the other training seed. Against scripted partners the same receivers reach 30.5 / 30.5\% (Table~\ref{tab:diagnosis}). The learned partners share one convention: the hand is fully extended between 0.93 and 1.20\,s after the start (5th--95th percentile over both seeds). The receiver reaches out immediately and takes the parcel on the fly, at about two-thirds of the partner's reach. Pool snapshots inherit the same timing, so the pool adds partner identities without adding partner behaviour.

\subsection{The main cause is partner start timing}
\label{sec:diagnosis}
We isolate the cause with paired inference-time interventions, changing one property of the partner at a time without retraining (Table~\ref{tab:diagnosis}).

\begin{table}[t]
\caption{\textbf{Causal diagnosis by intervention.} Supported transfer (\%) of the self-play receiver, seed 42 / seed 43, 200 scenarios per cell. One property of the partner is changed at a time. The own-partner baseline was re-evaluated for this table (87.0 against 89.5 in Section~\ref{sec:convention}).}
\label{tab:diagnosis}\centering\small\setlength{\tabcolsep}{6pt}
\begin{tabular}{lc}\toprule
Partner & Supported transfer\\\midrule
Own learned partner, unchanged & 95.0 / 87.0\\
\quad without its position corrections & 94.0 / 88.5\\
\quad without its tilt corrections & 95.0 / 89.0\\
\quad constant reach rate (cannot wait for the robot) & 93.5 / 81.5\\
\quad \textbf{start delayed by 1\,s} & \textbf{6.0 / 33.5}\\\midrule
Scripted partner, unchanged & 30.5 / 30.5\\
\quad start forced to the earliest (0.2\,s) & 57.5 / 51.0\\
\quad earliest start, fastest reach, no hesitation or sway & 80.0 / 65.0\\
\quad start forced to the latest (1.2\,s) & 27.0 / 19.5\\
\bottomrule\end{tabular}
\end{table}

Three results settle the question. First, removing the learned partner's position or tilt corrections changes nothing, and removing its ability to slow down costs a few points: the receiver's success does not rest on being helped. Second, delaying the receiver's own partner by one second, with everything else untouched, destroys the handover (95.0\% to 6.0\% and 87.0\% to 33.5\%). Third, the opposite intervention restores it: forcing a scripted partner to start early raises success from 30.5 / 30.5\% to 57.5 / 51.0\%, and a scripted partner made prompt, fast and smooth is handled in 80.0 / 65.0\% of episodes. Success falls monotonically with the delay (Figure~\ref{fig:repair}a).

\begin{figure}[t]
\centering
\includegraphics[width=0.9\linewidth]{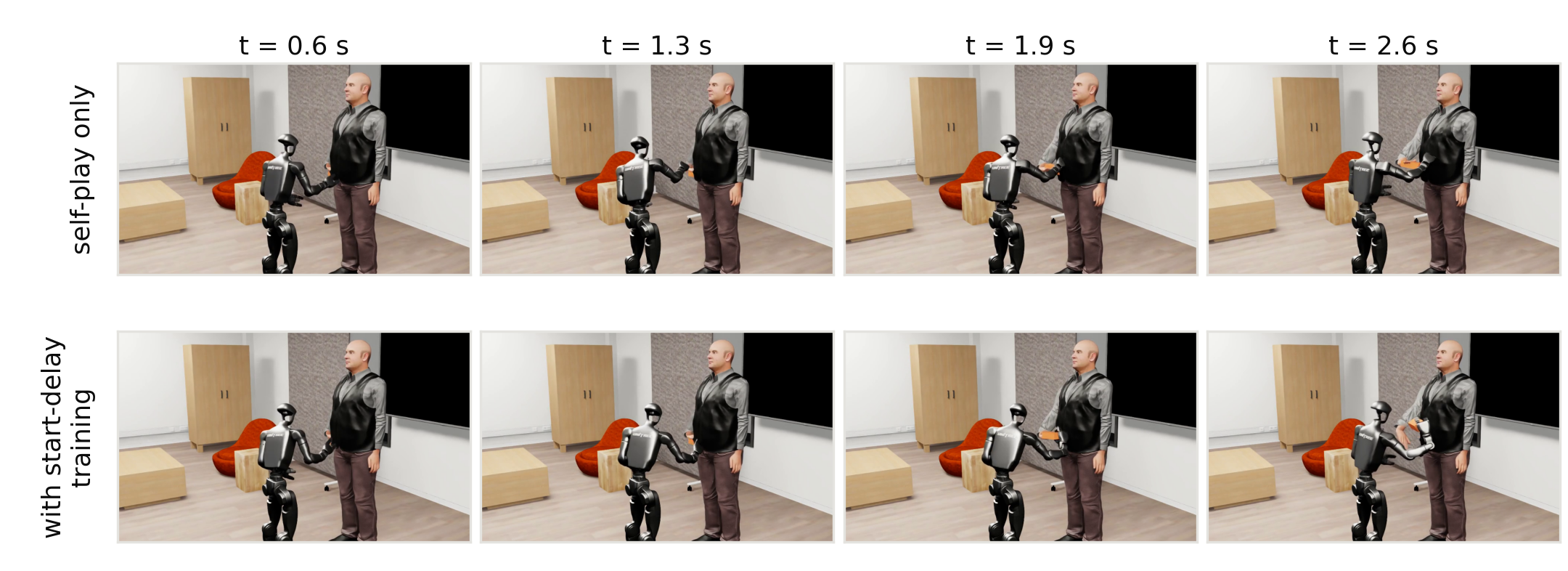}
\caption{\textbf{The timing failure and its repair, same scenario.} A scripted partner starts its offer about one second late. Top: the self-play receiver reaches at once, presses on the parcel as it comes up and never forms a grasp. Bottom: after \hsd{} training it takes the parcel as it arrives and completes a supported transfer. The backdrop is scenery only.}
\label{fig:frames}
\end{figure}

The mechanism is visible at the first touch (seed 42, 100 scenarios). With its own partner the receiver meets the parcel 9\,cm from the offer point with an open hand; its thumb presses on the parcel in 100\% of episodes and a stable grasp forms in 99\%. With the delayed partner it meets the parcel 30\,cm from the offer point with the hand already 42\% closed; the thumb gets around the parcel in 33\% and a stable grasp forms in 13\%. The receiver has learned where and when the parcel will be, not how to take it from a hand that arrives at another time (Figure~\ref{fig:frames}, top).

\subsection{Hidden start-delay randomisation repairs it}
Table~\ref{tab:repair} compares the matched continuations. \hsd{} removes the timing failure: with its own partner under random delay the receiver rises from 64.5 / 55.5\% to 94.5 / 88.5\%, and under a fixed one-second delay from 6.0 / 33.5\% to 93.5 / 89.0\%, while performance without delay is retained (95.5 / 93.0\%). The control, with exactly the same additional training but no delay, stays where it was (28.0 / 33.5\% under a one-second delay): the delay, not the extra training, is what repairs the policy.

\begin{table}[t]
\caption{\textbf{Effect of \hsd{}.} Supported transfer (\%), seed 42 / seed 43. All continuation arms start from the same self-play policy and receive the same 300 receiver updates. ``Own partner'' is the partner the receiver was last trained with. Fresh seed: never used for diagnosis, design or selection. ``--'': not evaluated.}
\label{tab:repair}\centering\small\setlength{\tabcolsep}{3.5pt}
\begin{tabular}{lcccc}\toprule
 & Self-play & Control & \hsd{} & \hsd{}, frozen\\
Partner & only & (no delay) & & partner\\\midrule
\multicolumn{5}{l}{\emph{Learned partner (development seed)}}\\
Own partner, random delay 0--1\,s & 64.5 / 55.5 & 67.5 / 59.5 & \textbf{94.5 / 88.5} & 97.0 / 91.5\\
Own partner, fixed 1\,s delay & 6.0 / 33.5 & 28.0 / 33.5 & \textbf{93.5 / 89.0} & 95.0 / 90.5\\
Own partner, no delay & 95.0 / 87.0 & 97.5 / 91.5 & 95.5 / 93.0 & 92.5 / 89.5\\\midrule
\multicolumn{5}{l}{\emph{Scripted partners, never trained on (development seed)}}\\
\texttt{train}, unchanged & 30.5 / 30.5 & 35.0 / 41.0 & \textbf{80.5 / 80.0} & 79.0 / 80.0\\
\texttt{train}, latest start & 27.0 / 19.5 & 12.0 / 33.5 & \textbf{83.5 / 72.0} & 78.0 / 78.5\\
Held-out bank, mean of 11 conditions & 30.5 / 27.2 & -- & \textbf{78.8 / 62.7} & 71.4 / 65.9\\\midrule
\multicolumn{5}{l}{\emph{Scripted partners (fresh seed)}}\\
Load family & 37.5 / 24.5 & -- & \textbf{85.0 / 80.0} & 85.0 / 79.5\\
Follow family & 24.5 / 21.0 & -- & \textbf{79.5 / 65.0} & 71.0 / 66.5\\
\texttt{train} & 37.5 / 24.5 & -- & \textbf{81.5 / 77.0} & 81.5 / 77.5\\
\texttt{late} (later than any training delay) & 23.5 / 11.0 & -- & \textbf{77.0 / 55.5} & 69.0 / 64.0\\
\bottomrule\end{tabular}
\end{table}

\begin{figure}[t]
\centering
\includegraphics[width=0.76\linewidth]{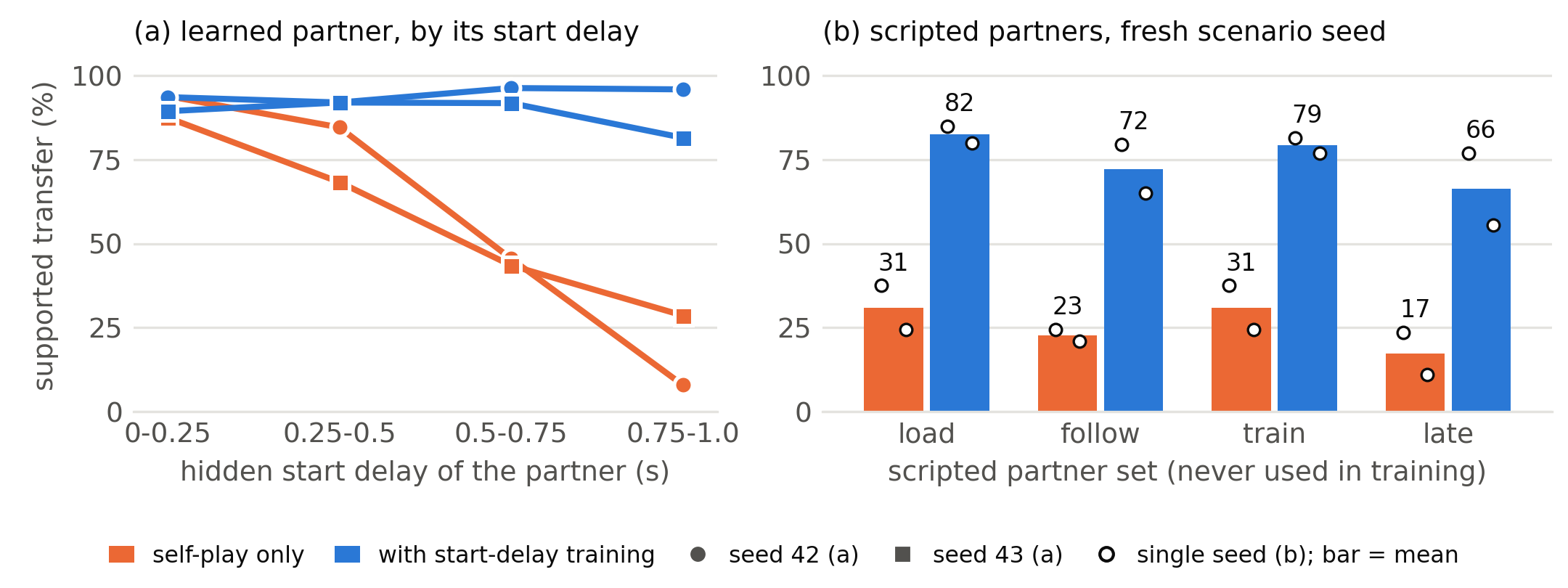}
\caption{\textbf{Start-delay training removes the timing dependence.} (a) Supported transfer with the receiver's own learned partner, by the partner's hidden start delay: self-play only collapses as the partner starts later; with \hsd{} it is flat. (b) Scripted partners on the fresh scenario seed; \texttt{late} starts later than any delay used in training.}
\label{fig:repair}
\end{figure}

\paragraph{The repair transfers to partners never seen in training.} With unchanged scripted partners the receiver rises from 30.5 / 30.5\% to 80.5 / 80.0\%, a 2.6-fold increase, against 35.0 / 41.0\% for the control, and it handles the latest-starting scripted partners (83.5 / 72.0\%) as well as early ones. The result replicates on the fresh scenario seed: 85.0 / 80.0\% with load partners, 79.5 / 65.0\% with follow partners and 77.0 / 55.5\% on \texttt{late}, whose start times lie beyond the training range (Figure~\ref{fig:repair}b). Across the eleven held-out conditions the mean rises from 30.5 / 27.2\% to 78.8 / 62.7\%.

\paragraph{The repaired handover is also gentler.} The share of held-out episodes with robot--partner contact above 20\,N falls from 10.7 / 7.9\% to 1.9 / 2.5\%: the receiver no longer waits in the path of the arriving parcel.

\paragraph{The repair needs no partner retraining.} Freezing the partner and continuing the receiver alone under the same delays gives the same robustness (last column of Table~\ref{tab:repair}): on the fresh-seed scripted sets the mean is 80.8 / 69.4\% with a learning partner and 76.6 / 71.9\% with a frozen one, a difference of $+4.1$ / $-2.5$ points, inside the five-point margin fixed before the run. \hsd{} can therefore be applied to any existing self-play pair at the cost of 300 receiver updates. It also helps when used from the start of training: trained with \hsd{} from scratch under the reference budget, the receiver reaches 81.0\% (seed 43) and 49.0\% (seed 42) with unchanged scripted partners, both above the 30.5\% of the reference; seed 42 entered the final curriculum stage late, at iteration 1040 of 1600.

\subsection{Holding still for the partner: the hidden release rule}
\label{sec:families}
How a partner lets go is a second hidden property, and the same recipe applies to it. Adding the hidden release rule to the \hsd{} policies teaches the receiver to satisfy partners that release only once it holds still: with its own partner in the settle style, supported transfer rises from 24.7 / 40.3\% to 79.2 / 88.3\%, while the load and follow styles stay near 90\%. The rule also improves transfer to unseen scripted partners, to 87.5 / 83.5\% (load) and 83.0 / 75.0\% (follow), up from 30.5 / 30.0\% and 24.5 / 20.0\% for self-play alone (Table~\ref{tab:families}).

\begin{table}[t]
\caption{\textbf{Release families.} Supported transfer (\%), seed 42 / seed 43, development seed. Top: scripted partners never trained on. Bottom: own learned partner under the release rule, by hidden style.}
\label{tab:families}\centering\small\setlength{\tabcolsep}{5pt}
\begin{tabular}{lccc}\toprule
 & Self-play only & \hsd{} & \hsd{} + release rule\\\midrule
Scripted load & 30.5 / 30.0 & 81.0 / 79.0 & \textbf{87.5 / 83.5}\\
Scripted follow & 24.5 / 20.0 & 70.0 / 66.0 & \textbf{83.0 / 75.0}\\
Scripted settle, with withdrawal & 4.0 / 3.5 & 2.0 / 0.0 & 3.5 / 1.5\\\midrule
Own partner, load / follow style & -- & 98.3 / 91.7,\; 87.3 / 79.4 & 88.3 / 93.3,\; 88.9 / 88.9\\
Own partner, settle style & -- & 24.7 / 40.3 & \textbf{79.2 / 88.3}\\
\bottomrule\end{tabular}
\end{table}

One partner type remains open. The scripted settle partner withdraws for good if the parcel moves more than 3\,cm before it lets go, and all receivers stay below 5\% on it. The receiver trained with the release rule does hold still, but only after an initial pull. Our diagnostics show that this test demands two behaviours at once, not taking the parcel from a hand that is still moving and holding still before pulling, which defines the next training target.

\section{Discussion and Limitations}
Because the partner shares the receiver's reward, cooperative self-play removes every source of variation that costs time, and the receiver's competence becomes conditional on a convention it was never asked to question. In handover this convention is temporal, it is dominated by one variable, and it is repaired by restoring the variation as a hidden constraint. Partner snapshots and sibling partners share the convention, so evaluating a self-play handover policy against them overstates its robustness by a factor of three. Since a frozen partner serves as well as a learning one, the repair is cheap: it requires only that the receiver experiences a partner whose start time it cannot predict.

\paragraph{Limitations.} The study is conducted in simulation, with a simulated partner hand and structured state instead of images, and with two training seeds. The scripted partner families are designed behaviours, and the training delay range overlaps their start range, which is why we also report \texttt{late} and the held-out bank.

\paragraph{Future work.} First, receivers that hold still \emph{before} pulling for partners that withdraw. Second, recorded human motion and people in the loop as held-out givers. Third, hidden rules for further physical variables, a budget-matched comparison with randomised scripted partners, and transfer to a physical arm with vision.

\section{Conclusion}
A self-play handover policy that is fluent with its own partner can be brittle to the simplest change in an unfamiliar one: when it starts to move. A one-second delay of the receiver's own partner reduces supported transfer from 95\% to 6\%. A hidden start delay on the learned partner repairs it, raising supported transfer with unseen scripted partners from 31\% to 80\% in both training seeds against 35--41\% for a matched control, with consistent gains on a fresh scenario seed and a held-out bank and with no retraining of the partner. Hidden, enforced partner variation turns cooperative self-play from a fluent pair into a receiver that can be handed an object by someone else.

\bibliography{references}
\end{document}